\documentclass[10pt]{article}
\newif\ifblindreview

\usepackage[letterpaper]{geometry}
\usepackage{amta2024}
\usepackage{times}
\usepackage{url}
\usepackage{latexsym}
\usepackage{natbib}
\usepackage{layout}
\usepackage{multicol}
\usepackage{booktabs,array}
\usepackage{float}
\usepackage[hidelinks]{hyperref}
\usepackage[all]{hypcap}
\usepackage{graphicx}
\usepackage{inconsolata}
\usepackage{xcolor}
\hypersetup{
    colorlinks,
    linkcolor={red!50!black},
    citecolor={blue!50!black},
    urlcolor={blue!80!black}
}

\usepackage{tabularx}
\usepackage{soul}
\usepackage{comment}
\usepackage{amsmath}

\ifblindreview
  \newcommand{\authorinfo}{\author{}} 
\else
  \newcommand{\authorinfo}{
    \author{\name{\bf Arle Lommel} \hfill \addr{arle.lommel@gmail.com}\\
    \addr{\small CSA Research, Massachusetts, United States}\\
    \name{\bf Serge Gladkoff $^{**}$} \hfill   
    \addr{serge.gladkoff@logrusglobal.com}\\
                \addr{\small Logrus Global LLC, Pennsylvania, United States}\\
    \name{\bf Alan Melby} \hfill \addr{melbyak@yahoo.com}\\
    \addr{\small Professor Emeritus of Linguistics at Brigham Young University, United States}\\
    \name{\bf Sue Ellen Wright} \hfill \addr{swright@kent.edu}\\
    \addr{\small Professor Emerita, Kent State University, United States}\\
    \name{\bf Ingemar Strandvik $^*$} \hfill \addr{ingemar.strandvik@ec.europa.eu}\\
    \addr{\small Directorate-General for Translation, European Commission, Belgium}\\
    \name{\bf Katerina Gasova} \hfill   
    \addr{katerina.gasova@gmail.com}\\
    \addr{\small Global Quality Solution Strategist, Argos Multilingual, Czechia}\\
    \name{\bf Angelika Vaasa} \hfill   
    \addr{angelika.vaasa@europarl.europa.eu}\\
    \addr{\small Directorate-General for Translation of the European Parliament, Luxembourg}\\
    \name{\bf Andy Benzo} \hfill   
    \addr{andybenzo@jurismentis.com}\\
    \addr{\small American Translator's Association President-Elect, 2024}\\
    \name{\bf Romina Marazzato Sparano} \hfill   
    \addr{romina@languagecompass.com}\\
    \addr{\small ISO WG11 TC 37 Plain Language Standard Project}\\
    \name{\bf Monica Foresi} \hfill   
    \addr{fmonica07@gmail.com}\\
    \name{\bf Johani Innis} \hfill 
    \addr{grabavac@icloud.com}\\
            \name{\bf Lifeng Han $^{**}$ \and Goran Nenadic} \hfill \addr{lifeng.han, g.nenadic@manchester.ac.uk}\\
            \addr{\small The University of Manchester, United Kingdom}
    \\  \textit{$^*$: the opionions expressed are the author's alone and do not represent the European Commission's official position. $^{**}$: corresponding authors}
    } 
  }
\fi

\begin{document}

\amtaHeader{x}{x}{xxx-xxx}{2015}{45-character paper description goes here}{Author(s) initials and last name go here}
\title{\bf The Multi-Range Theory of Translation Quality Measurement: MQM scoring models and Statistical Quality Control}
\authorinfo

\maketitle
\pagestyle{empty}

\begin{abstract}
\vspace{5pt}
  The year 2024 marks the 10th anniversary of the Multidimensional Quality Metrics (MQM) framework for analytic translation quality evaluation. The MQM error typology has been widely used by practitioners in the translation and localization industry and has served as the basis for many derivative projects. The annual Conference on Machine Translation (WMT) shared tasks on both human and automatic translation quality evaluations used the MQM error typology.
  The metric stands on two pillars: \textit{error typology} and the \textit{scoring model}. The scoring model calculates the quality score from annotation data, detailing how to convert error type and severity counts into numeric scores to determine if the content meets specifications. Previously, only the raw scoring model had been published. This April, the MQM Council published the \textit{Linear Calibrated Scoring Model}, officially presented herein, along with the \textit{Non-Linear Scoring Model}, which had not been published before.
  This paper details the latest MQM developments and presents a universal approach to \textbf{translation quality measurement} across three sample size ranges. It also explains why Statistical Quality Control should be used for very small sample sizes, starting from a single sentence. 

\end{abstract}

\begin{multicols}{2}

\section{Introduction and Background}

Machine Translation (MT) was one of the earliest artificial intelligence (AI) tasks when Machine and Intelligence was launched in the 1950s in the wake of WWII \citep{han2022thesis_MWE_MT}. 
MT has significantly influenced the translation industry since the statistical MT (SMT) models started to produce editable automatic translations in the early 2010s, just before neural MT (NMT) came to the stage in the middle and second half-decade \citep{koehn2009statistical,DBLP:journals/corr/BahdanauCB14,google2017attention}. But even today, when Generative AI (GenAI) has captured the imagination of billions of people, both human and AI-based translations may still contain errors.

Translation errors often carry risks. Their consequences range from minor misunderstandings to serious legal, financial, reputational, or health-related outcomes for end users, translation providers, clients, and other stakeholders \citep{han2024neural}. Risk mitigation requires evaluation to identify and quantify these risks.
The Multidimensional Quality Metrics (MQM) framework for analytic Translation Quality Evaluation (TQE) was first proposed by \cite{MQM2014} just before the arrival of NMT, originally published as a deliverable of the EU-funded QTLaunchpad project. From the very beginning, it was designed for evaluation of both human translation (HT), and machine translation (MT), and it can now be applied to AI-generated translation.
MQM has formalized and standardized the so-called analytic approach to translation quality measurement. 

This approach is typically based on evaluating a sample. It involves annotating translation errors by attributing them to predefined \textbf{error types} and \textbf{severity levels} to generate the data for deriving the translation quality score. 
MQM appeared as a major and fundamental standardization attempt to alleviate the then-widespread problems of practical translation evaluation, at a time when there was no single way to approach translation quality measurement.
However, the lack of a sophisticated design of hierarchies and an adaptable scoring model also posed a bottleneck for its real-world application.

The original EU-funded QTLaunchpad project deliverable, the MQM 1.0, published on the W3C project page, included only the raw scoring model, in which the score is calculated as a direct proportion of errors found in the evaluated sample, to the size of the sample.
This approach has several drawbacks, specifically, a) such scores do not use human-readable scales, b) they have varying and non-intuitive error tolerance thresholds, and c) they produce non-comparable quality values across various content types and scenarios.
All other score calculation models also had fixed, non-adaptable scoring systems, which confused the industry and led to numerous `reinventions of the wheel'.

Subsequently, a few other human-centric evaluation metrics were proposed with a similar approach to MQM and these efforts were more simplified and easy to deploy, such as the HOPE metric by \cite{gladkoff-han-2022-hope}, which only includes eight initial error types and error severity levels. This approach was refined from industrial practice and designed specifically for machine-translation outputs. It also featured very different scoring models.

Nevertheless, the MQM framework has been picked up again by the leading MT shared task venue WMT since 2021 as the official human evaluation strategy to judge the submitted MT systems \citep{freitag-etal-2021-results,freitag-etal-2022-results,freitag-etal-2023-results}.
There have been automatic evaluation designs that aim to mimic the MQM idea, such as COMET-MQM reported from WMT2020 metrics shared task \citep{mathur-etal-2020-results}.

In this paper, we introduce the latest developments of the MQM framework from the MQM Council which comprises a voluntary, community-driven research and standardization group composed of experts interested in translation quality evaluation, who have been developing the MQM since 2016. 
\footnote{\url{https://themqm.org/mqm-council/}} 
These start with the discussion of \textbf{Sampling} and \textbf{Low IRR} phenomena, followed with \textbf{sample-sizes}, \textbf{MQM2}, Projecting PI and \textbf{Formulas}, and the introducing \textbf{non-linearity}. We leave the detailed \textit{MQM Parameters} into appendix for indexing.

    In recent years, the widely used DQF subset of MQM has been improved and updated to become MQM Core. This error typology is better adapted to quality management systems with a clearer structure for devising improvement actions.
The latest iteration of the framework includes the revised MQM Core and MQM Full error typology, a new linear scoring model with calibration, a process description with a sample scorecard, and now a non-linear MQM scoring model. (appendix \ref{sec:non-linear-mqm}).

We further argue that different evaluation approaches have to be used for three ranges of sample sizes. For the first time in the history of the translation and localization industry, to the best of our knowledge, this paper presents a multi-range, versatile theory and technique of Translation Quality Evaluation, making it possible for interested researchers to construct almost any analytic metrics derived from this approach.
We also explained in this paper why segment-level scores cannot be accurate in principle, and explain the area of applicability of Statistical Quality Control to Translation Quality Evaluation.

\section{On Sampling}
\label{sec:sota-MQM}

For any statistical approach to be applicable, it is critical to know what statistical distribution is valid before choosing the right distribution for further statistical analysis. Such analysis is based on the notion that, \textit{ideally}, (a) all errors are independent and (b) the probability of errors in the text is uniform.
Although these assumptions are always made, \textit{neither} assumption is true concerning texts (and translation products) in general.

Practitioners in the language and translation industry know that translation errors are not, in general, uniformly distributed in content and, what is more, over time. Furthermore, their significance and “weights” are also different in various parts of the material and/or types of material, and also vary according to other sometimes unpredictable factors. In addition, different types of errors may depend on each other. Indeed, some errors only occur when triggered by other fundamental errors.

There are about a dozen very important factors that can influence why error distribution in text is not, in fact, uniform.
Because of this, it is always recommended to revise, review, and evaluate the entire text.
But no one has the resources to fully evaluate everything. That's why translation quality evaluation is typically based on evaluating samples (such as the work by \cite{gladkoff-etal-2022-measuring} from an industrial setting).
Sometimes this sampling is done by selecting full-text samples from a larger population, e.g. evaluating every tenth translation fully, or selecting complete shorter texts to arrive at the determined sample size.
A typical approach in a traditional localization process is to select samples of medium sizes (500-5000 words) and then apply an analytical approach.
However, smaller samples and larger samples are also possible, and it is important to consider the entire range of possible sample sizes, from one sentence to very large documents, which is the purpose of this paper.

\section{Low IRR is not a bug, but a feature}
\label{sec:sota-MQM}

Human translation quality evaluations on small samples have low Inter-Rater Reliability (IRR). What matters in human evaluation is that trained linguists tend to agree (demonstrate high IRR) on bigger samples, but on a segment level two linguists often disagree
on the same issue. 
This phenomenon has been widely recognized by both experts in the translation industry and data science \citep{gladkoff2023student}.
Research in the field of translation error annotation has demonstrated that linguists can disagree even on the precise scope of errors, as well as about error categorizations and severity attributions \citep{lommel2014assessing}. 
The researchers established that these issues are not a consequence of low linguist qualifications or technical problems.
Multiple factors at play can determine disagreement among annotators: 1) the complexity of the task; 2) the ambiguity of the text; 3) the quality of the translation; 4) subjectivity; etc.

Text is not the information itself; it is merely the conveyor of the expression of intended meaning. Moreover, not only can we not know for sure the author's intended meaning, which leaves the text open to the reader's interpretation. This interpretation depends on the reader's cultural, educational, and professional background and experiences, as well as on their skills and the context of the communication act.
Additionally, language is highly ambiguous, offering numerous expression tools that allow a single sentence to be interpreted in many ways. This inherent ambiguity leads to significant uncertainty in any error annotation, which is a fundamental property of language rather than a flaw of human assessors \citep{gladkoff-etal-2022-measuring}.

Data scientists often make shallow conclusions from low IRR, believing that automatic calculation of errors will resolve this problem.
In recent years, many automatic metrics have been constructed, often claiming ``human judgment" quality. Recently, AI-based metrics have appeared, along with unverified and unproven claims that ``GenAI seems capable of measuring translation quality" \citep{gladkoff2024mtuncertaintyassessingneedpostediting}.
In reality, however, GenAI does neither ``think" nor ``understand" anything and for this reason, the factual accuracy of GenAI generation remains a huge problem.
No language model is Turing complete \footnote{\url{https://ai-lab.logrusglobal.com/why-no-agi-can-be-built-with-language-models/}}. Language models are constrained by their architecture and the limitations of their training data and computational resources. Language models are specialized tools designed for processing and generating text based on patterns learned from vast amounts of data. They are good at some tasks such as language generation, and summarization, but they are not capable of performing arbitrary computations or reasoning in the same way that a Turing complete system can. Because of this, language models are not considered general-purpose reasoners. They can provide useful responses and assist with many tasks within their scope of capabilities, but they cannot reason and compute in the general-purpose manner that a Turing machine or Turing complete system can. Among other things, the factual accuracy of GenAI generation remains to be a huge problem. This affects both error annotation capability and accuracy on a secondary process level. GenAI may produce fluent text, but it performs worse on derivative tasks, such as error annotation tasks. In other words, GenAI misses factual errors when generating content, and of course, does not ``see" errors during the annotation process, which also results in issues with the accuracy of this process.
Additionally, the better the GenAI output is, the more variable it becomes, which is similar (although different in nature) to the variability in human judgment.
Human judgment is variable because different people may have their own interpretation of the text. Advanced GenAI behaves similarly, with its variable output becoming another interpretation. However, this interpretation is not verified or supported by human intelligence.

And similar to human evaluation, the smaller the context window of the text, the more variable is the GenAI response.
For all these important reasons, the smaller the evaluation sample, the greater the uncertainty of translation quality annotation. This is due to the intrinsic, fundamental variability in both human and GenAI-based error annotation that leads to the uncertainty of error evaluation.
This means one simple thing: at the low end of the scale, sentence-level automatic scoring is so unreliable that it makes no sense, regardless of the method used to produce the score.

Furthermore, all automatic metrics must be supported by proper benchmarking and validation, which takes a lot of time and is very specific to the particular implementation setting and the specifications (language pair, subject matter area, task requirements, etc.).

Automatic metrics of any kind produce a single number with unknown reliability and confidence intervals. By definition, this number ignores the ambiguity of the text and therefore disregards other valid interpretations, which can be validated by human evaluators on larger samples.

For all we know, various automatic and GenAI measurement results must be validated by analytic human evaluation on samples of sufficient sizes, which converges to higher IRR (inter-rater reliability) in controlled settings with training and monitoring \citep{gladkoff-etal-2022-measuring,han2024neural}.

Training and continuous validation of evaluators' work contribute to improving IRR over time. What ultimately matters for reliable evaluation is that a proper process allows for achieving statistically valid IRR of human evaluation at the sample level, not necessarily at the segment level.

\section{Three sample-size ranges -- three very different methods}
\label{sec:sota-MQM}

The industrial era has developed a vast mathematical and methodological apparatus for measuring the quality of products by evaluating small samples from very large production lots, e.g. using \textit{student's t-distribution} \citep{student1908probable} such as the recent work by \cite{gladkoff2023student}.
In a setting where decisions about the quality of a large lot are based on small samples, the uncertainty is so high that only sophisticated methods of statistical quality control can handle such a problem. They have been extensively developed and described by many researchers, such as \cite{montgomery2019introduction}, and have been long standardized by ISO \footnote{ISO 2859-2:2020 Sampling procedures for inspection by attributes \url{https://www.iso.org/standard/64505.html}}.
In a nutshell, errors \textbf{\textit{always}} have a statistical nature, and this is yet another reason why segment-level quality scores do not make sense -- not only the methods of producing them must be properly benchmarked and verified, but the fundamental uncertainty of individual error annotation is so high that it is not possible to get a ``true" measurement except for very trivial mechanical spelling errors.

When measuring errors on a very small sample, we need to apply the Statistical Quality Control (SQC) method.
These methods are extremely complex due to the underlying statistical and mathematical apparatus, which is why ISO 2859-1 consists of 94 pages of tables.
Additionally, the conclusions drawn from applying these methods are not quality ratings but rather probabilities of the producer's and consumer's risks, which can be determined using such methods.

In an earlier work by \cite{gladkoff-etal-2022-measuring}, 
it was demonstrated that for the translation quality rating to have low variability, the sample size should be greater than 200 sentences (3400 words). The authors discuss that if they go to smaller sample sizes, the confidence interval explodes.
This means that even a 2000-word sample introduces significant uncertainty in the quality measurement result, and even more so with samples of 1000 words and especially \textit{500 words}.

The lowest known credible metric is the ATA certification model \citep{han2022meta}, where linguists are asked to translate one page (250 words). To address the uncertainty of evaluating such a small sample, two different reviewers assess the work, effectively duplicating the evaluation effort to mitigate the risks associated with the limited sample size. Additionally, the ATA certification requires linguists to translate two one-page samples to see how they handle the translation of different types of text, which also helps reduce the uncertainty of evaluating small samples.

The methods of Statistical Quality Control (SQC) are outside the scope of this paper. However, for this article, it is important to note that the translation quality of a sample smaller than 15-17 sentences (one page, 250 words) falls into the realm of SQC and cannot be measured by analytic quality evaluation methods unless the sample covers the entire text.

For samples of approximately 300 words and above, the effects of statistical uncertainty and low Inter-Rater Reliability cease to have a significant adverse effect, making methods of analytical quality evaluation applicable. For samples larger than approximately \textit{5000 words}, other effects start to manifest, such as the priming effect of human perception. Non-linearity starts to become apparent with larger samples. In this paper, we introduce a non-linear calibration model that works for samples of both medium (customary) and large sizes.
The three ranges of sample sizes are governed by entirely different mathematical apparatus. These ranges are shown in Figure \ref{fig:three-ranges-of-tolerance} (word counts: 0, 500, 5000+).

\section{MQM 2.0: The State of the Art of Analytical Quality Evaluation}
\label{sec:sota-MQM}

MQM is a framework for analytic Translation Quality Evaluation (TQE). It can be used to evaluate human translation (HT), machine translation (MT), or AI-generated translation. MQM consists of two key components: the error typology and the scoring model. The MQM error typology is organized hierarchically with \textit{seven} high-level error dimensions, subordinate error types, and associated severity levels. The \textit{scoring model} features a system of weights and parameters assigned to the error types and severity levels, as well as a scoring formula used to calculate a numerical score that represents the quality of the evaluated translation according to agreed-upon specifications.

The evaluated sample can comprise an entire document or a set of documents, or parts thereof. Evaluators frequently work with samples in the \textit{range} of 500 to 20,000 words, depending on the size of the project and the resources available for evaluation.
%

\subsection{Error Typology}
As noted above, the MQM error typology is based on seven high-level dimensions, with subordinate error subtypes at various hierarchical levels. For example, the Accuracy error dimension contains error subtypes such as Addition, Mistranslation, and Omission. At the next hierarchical level, Mistranslation, for example, contains error subtypes such as Misrepresentation of technical relationship, False friend, MT hallucination, etc. The complete repository of all error types is known as \textbf{MQM-Full}. Implementers typically do not use the complete repository but select a subset of MQM-Full to provide the granularity they need for their implementation context.

\textbf{MQM-Core} is a pre-defined subset that comprises the seven high-level error dimensions with the selected error sub-types that are most widely used in the language sector.
The error types are represented by names and their rigorous description. They have a specific, defined meaning and should not be understood as general language words or common terms. For instance, Accuracy in MQM refers to the appropriate correspondence between the source and target language, rather than to factual correctness in general.

\subsection{Scoring Model}
The second key component is a scoring model. The scoring model is a method, process, and formula for deriving quality scores resulting from error annotation data based on customer specifications.

Implementers design their scoring model by selecting error dimensions and sub-types with the granularity relevant to the implementation environment. Implementers assign penalty points or error weights to the error types and define penalty multipliers for the severity levels. Thereafter, for each identified error instance, evaluators assign an error type and severity level, and record them in the translation environment or on a scorecard. These values are then used by the scoring formula to calculate a \textbf{Quality Score}.
The scoring calculation determines the final Quality Score. In this paper, we distinguish three types of scoring model:

\begin{itemize}
    \item Linear Raw Scoring Model
    \item Linear Scoring Model with Calibration
    \item Non-Linear Scoring Model with Calibration
\end{itemize}

Linear Scoring allows the calculation of a quality score with or without calibration. The Raw Linear score calculates the portion of the evaluated text with errors and subtracts this value from 100 to get the value which directly represents the error-free portion of the evaluated sample.

In April 2024, the MQM Council published the extended MQM 2.0 scoring model document, which includes the Linear Scoring Model with Calibration \footnote{MQM extended scoring model document \url{https://themqm.org/error-types-2/the-mqm-scoring-models/}}. The idea of calibration is to set the \textbf{Passing Interval} using a separate, special score scale for the convenience of human use of scores.

The two scoring methods -- with and without calibration –- serve different purposes.
Non-calibrated scores represent the raw results of an evaluation task and are easy to calculate, but difficult to interpret and use; in addition, different tolerance thresholds are not intuitively represented on a raw score scale with varying positions and non-integer numbers, which are different for different scenarios.

Calibrated scores are more complex to produce, but are convenient for humans to use, and also enable implementers to create scoring models that are comparable across various content types, use cases and service levels. For example, a translation service provider is likely to use different calibrations for different clients or use cases, but the resulting calibrated score scale will be the same, making it easy to work with for all stakeholders in all scenarios. With a little additional effort for calibration, the quality scores can be made much more universal, human-readable and useful.

Linear scoring models apply the same scoring irrespective of the sample length. However, human perception of the quality tends to be different depending on the size of the sample. The \textbf{Non-Linear Scoring Model} takes into account changing human perception throughout content consumption and produces accurate scores across a wide range of sample sizes, from small ones to infinity. These factors are explained further in this paper.

The \textbf{Raw Scoring Model (Score without Calibration)} works with the basic values and parameters: the evaluation word count and the total of the penalty points calculated for the sample, as defined below.

The \textbf{Scoring Model with Calibration} works with all the same original values and parameters as the Raw Scoring Model, but a few additional parameters are required as explained in Score Calculation with Calibration (\ref{subsec:calibration}).

A Pass or Fail rating is assigned about the established passing threshold or error tolerance value. With calibration, setting a relevant quality threshold and error tolerance limit is much easier and more flexible, making the pass/fail decision clearer and more understandable. In addition, calibration allows for the adjustment of the scoring formula to match the perception of the rater.

\begin{figure*}[t]
\begin{center}
\includegraphics[width=1\textwidth]{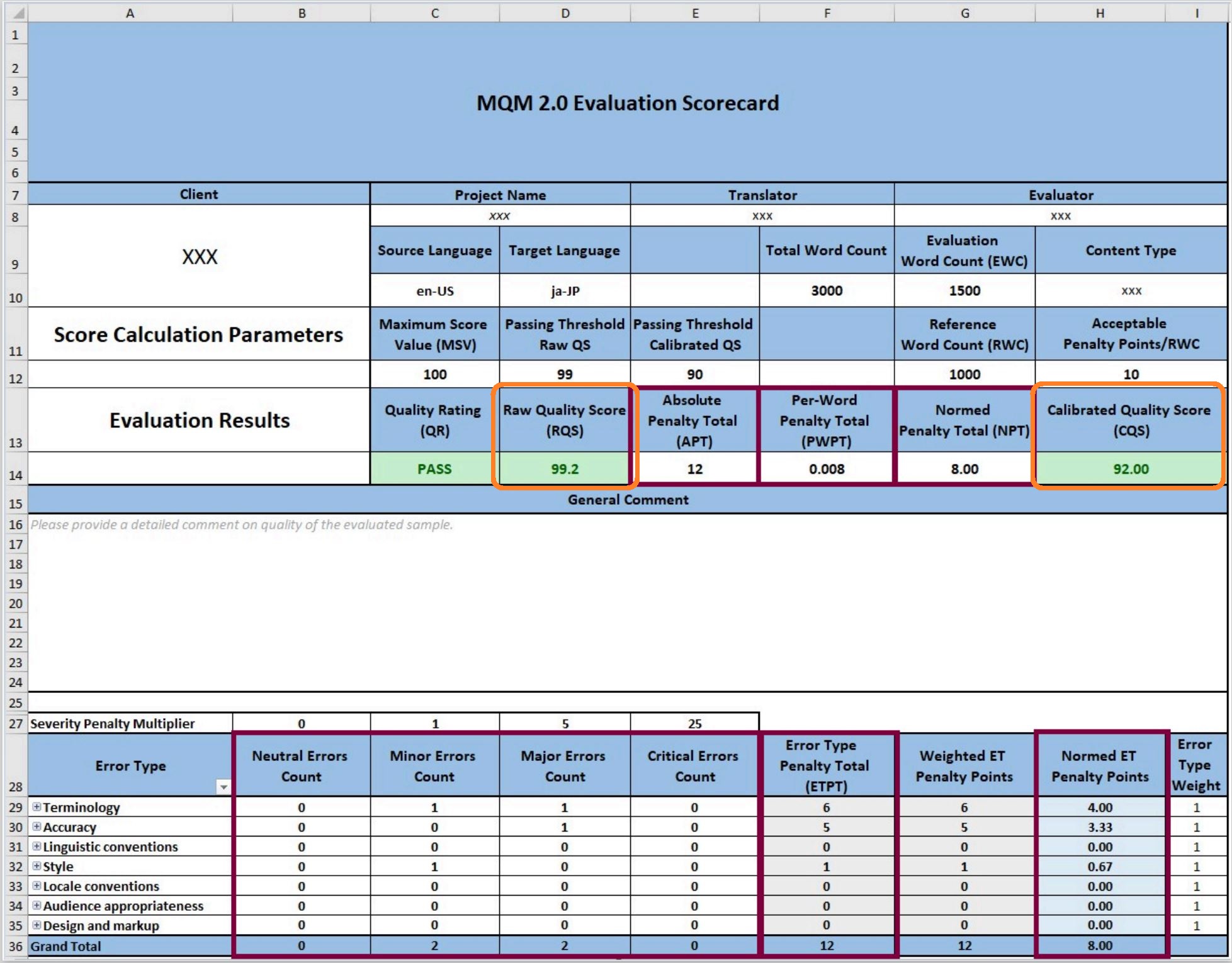} 
    \caption{MQM2.0 Evaluation Scorecard: Quality Measure and Tools (red squares with straight corners), and Quality Scores (orange squares with curved corners). }
    \label{fig:mqm2.0-scorecard}
\end{center}
\end{figure*}

\subsection{Setting up an MQM evaluation system}
To set up the MQM evaluation process, implementers must first create a specific metric by:
    1) Evaluating translation project specifications and analyzing them concerning end-user needs;
    2) Selecting Error Types from the MQM Error Typology that are appropriate to the defined specifications to create the error typology for the metric;
    3) Integrating the created MQM metric, suitable for their specific use case, into a scoring system into their tool environment or using it in a scorecard format;
    4) Determining sample size ranges and defining sampling procedures.

Historically, translation evaluators have used spreadsheet-like scorecards (see Figure \ref{fig:mqm2.0-scorecard}), but today evaluators are more likely to annotate errors and integrate scoring calculations in specialized tools or utilities built into their translation environment. To visualize and describe the annotation and score-generation process, it is helpful to use a scorecard as an example. The sample scorecards used here represent useful options, but they are not normative.

The default setting for MQM is to use four severity levels, with the severity multipliers 0-1-5-25. Implementers can choose to use different ranges of severity levels, as explained later (\ref{subsub:penaltyMultiplier}), and implement them via their respective Severity Penalty Multiplier. (Appendix \ref{sec:metric-n-quality-model-app} for detailed parameters)

\section{Projecting PI and the Formulas}
\begin{figure}[H]
\includegraphics[width=0.49\textwidth]{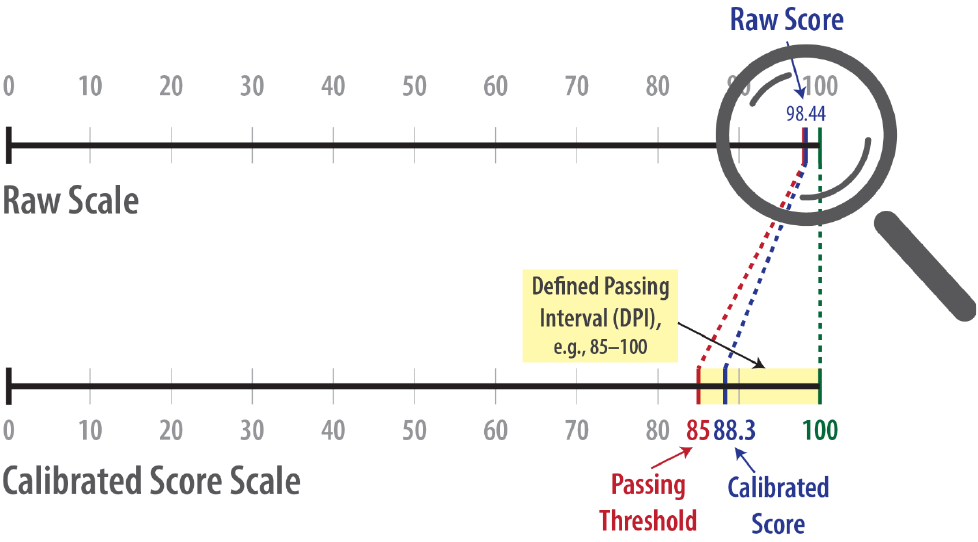} 
    \caption{Projecting the small Passing Interval window in the Raw-Score scale to the scale of the Calibrated Score, where the Passing Threshold is chosen arbitrarily by the customer based on the relevant values that apply to a specific context.}
    \label{fig:mqm2.0-scorecard-scaling-factor}
\end{figure}

\begin{figure*}[t]
\begin{center}
\includegraphics[width=0.9\textwidth]{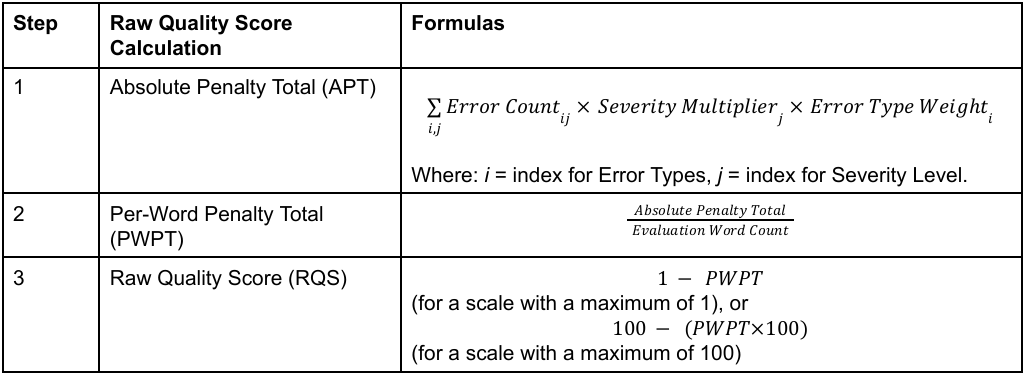} 
    \caption{Formulas for raw score calculation. }
    \label{fig:mqm2.0-scorecard-formulas}
\end{center}
\end{figure*}

Figure \ref{fig:mqm2.0-scorecard-scaling-factor} displays the projecting of the small passing interval window in the raw-score score to the scale of the calibrated score. Figure \ref{fig:mqm2.0-scorecard-formulas} introduces the formulas for raw score calculation. 

\begin{figure*}[]
\begin{center}
\includegraphics[width=0.9\textwidth]{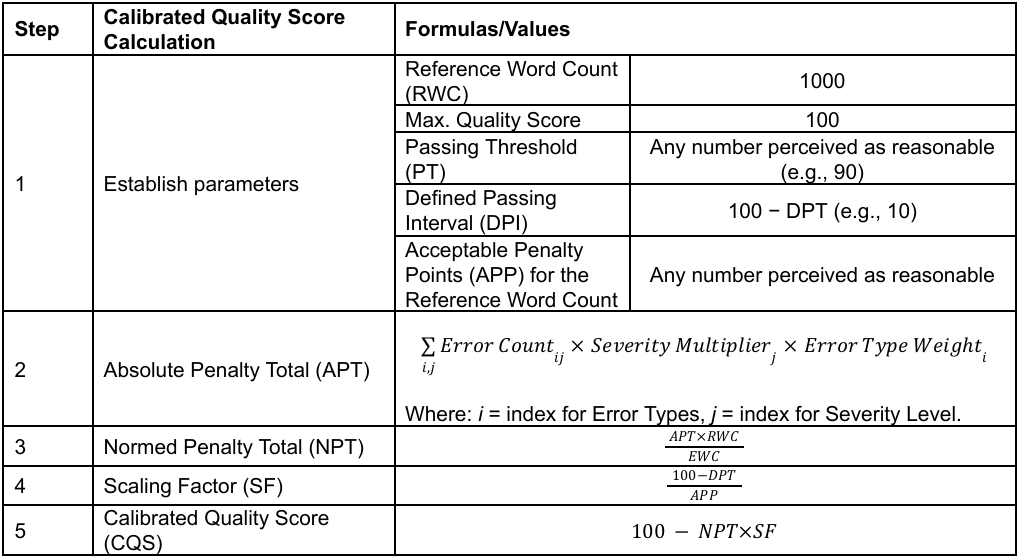} 
    \caption{Formulas for calibrated score calculation.}
    \label{fig:mqm2.0-scorecard-formula-calibration}
\end{center}
\end{figure*}

\section{On Non-Linearity}
\label{sec:non-linear-mqm}
This section describes the non-linear scoring model which solves the problem of the non-linearity of human perception concerning samples of medium to large size.

All is well with linear scoring until the sample size is either too small (requiring methods of Statistical Quality Control) or too large. Many practitioners have observed that linear scoring defined for small to medium samples does not work for larger chunks of text.
The reason that changing human perception fails a linear model is that the human mind is not a fixed state machine – we change our behaviour and our opinions with experience.
This effect is known in cognitive science as “priming,” where encountering certain features in the text makes readers more alert to them \textit{\textbf{as they work their way through the text}}.

Priming works on errors too. When we see two errors on one page, then two errors on the next page, and three errors on the page thereafter, we get the impression that there are simply too many errors in the document.
This is observed in real translation quality evaluation practice as follows (this is an actual quote from a very large buyer of translation services):
Once we started using our current methodology in 2020, we still asked the evaluators to indicate the cases where their actual feeling was different from what the score gave them. We very quickly realized that the main issue was that with very short samples the scoring was overly harsh and with very long samples it was too lenient. The reason for this is that when we evaluate holistically, the perception is not congruent with our scoring formula. For example, we might feel that if a translation sample is about one page, a single major mistranslation error is enough to say that it fails. However, if the sample is seven pages, we are not okay with allowing seven major mistranslation errors before it fails. Instead, we would prefer to fail the sample already at three or four errors.
This poses a problem for the linear scoring model which simply prorates the number of errors per page to a total number of pages in the sample."

\subsection{Non-Linear error tolerance – what it may look like?}

\begin{figure*}[t]
\begin{center}
\includegraphics[width=0.9\textwidth]{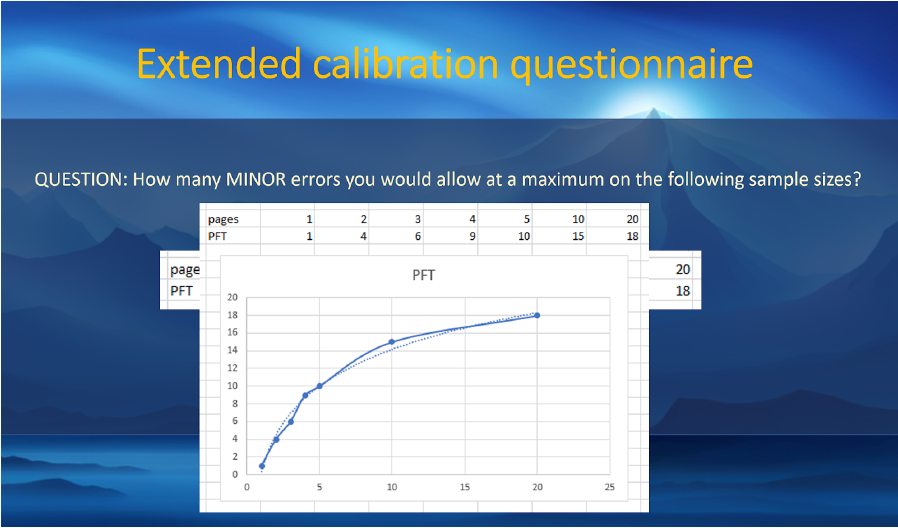} 
    \caption{Extended calibration questionnaire. }
    \label{fig:extend-calibration-questionaire}
\end{center}
\end{figure*}

A standard calibration questionnaire only asks how many minor errors are acceptable/not acceptable on the standard sample size.
You only need one data point to draw the straight line which originates from the zero point.
But if we believe that the quality tolerance is non-linear, we need more error points to see what the curve might look like.
We have made numerous surveys of the quality specialists with an extended calibration questionnaire which asks for a tolerance threshold for several sample sizes.
All of them follow the same basic pattern: the error tolerance quite sharply decreases with increasing size of the sample, as shown in Figure \ref{fig:extend-calibration-questionaire}.

\begin{figure*}[t]
\begin{center}
\includegraphics[width=0.9\textwidth]{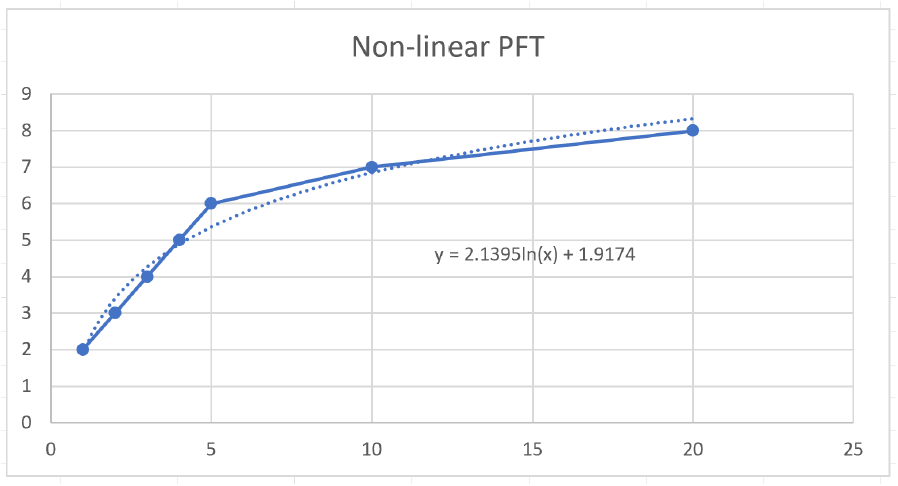} 
    \caption{Real world non-linear calibration questionnaire with parameters found. }
    \label{fig:real-world-nonlinear-calibration-questionaire-cropped}
\end{center}
\end{figure*}

Of course, it is difficult for quality managers to answer such questions, because they are trying to calculate the number based on the linear model, so in order to respond to this survey correctly, it’s best: to either ask the quality manager who is not so proficient with the linear scoring formula, or specifically ask NOT to prorate the tolerance based on a standard sample.
With this complication in mind, and properly taken into account, all the calibration surveys end up with one result as shown on Fig \ref{fig:extend-calibration-questionaire}.
This is a logarithmic dependency which can be easily calculated in Excel as a logarithmic trend line with concrete parameters.

Naturally, the data points for this curve are all empirical, but this is a strength, not a weakness, of the calibration approach. What non-linear calibration does is capture the reality of the non-linearity of human perception and extend the applicability of the MQM Scoring Model.

\subsection{Benefits of non-linear scoring: faithfulness to human perception and scalable to a wide range of sample sizes}

Now, as we found what error tolerance looks like for a wider range of sample sizes, we understand that the linear model only works on a very small range of sample sizes near the standard sample size that the model has been developed for.
If the standard sample size is 2000 words, then already the metric won’t work correctly for 1000 and 5000 words!
This is illustrated on the chart below, where you can easily see that a linear scoring formula snapped to just one “standard” tangent point will be very far from actual human perception on a very different sample size.

\begin{figure*}[t]
\begin{center}
\includegraphics[width=0.9\textwidth]{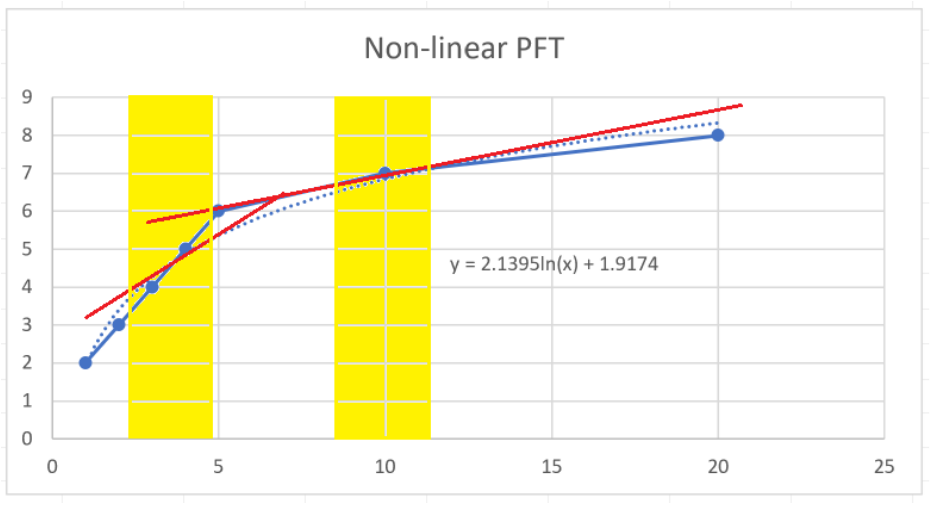} 
    \caption{Linear scoring formula snapped to just one “standard” tangent point will be very far from actual human perception on other sample sizes. }
    \label{fig:linear-vs-nonlinear}
\end{center}
\end{figure*}

\begin{figure*}[t]
\begin{center}
\includegraphics[width=0.9\textwidth]{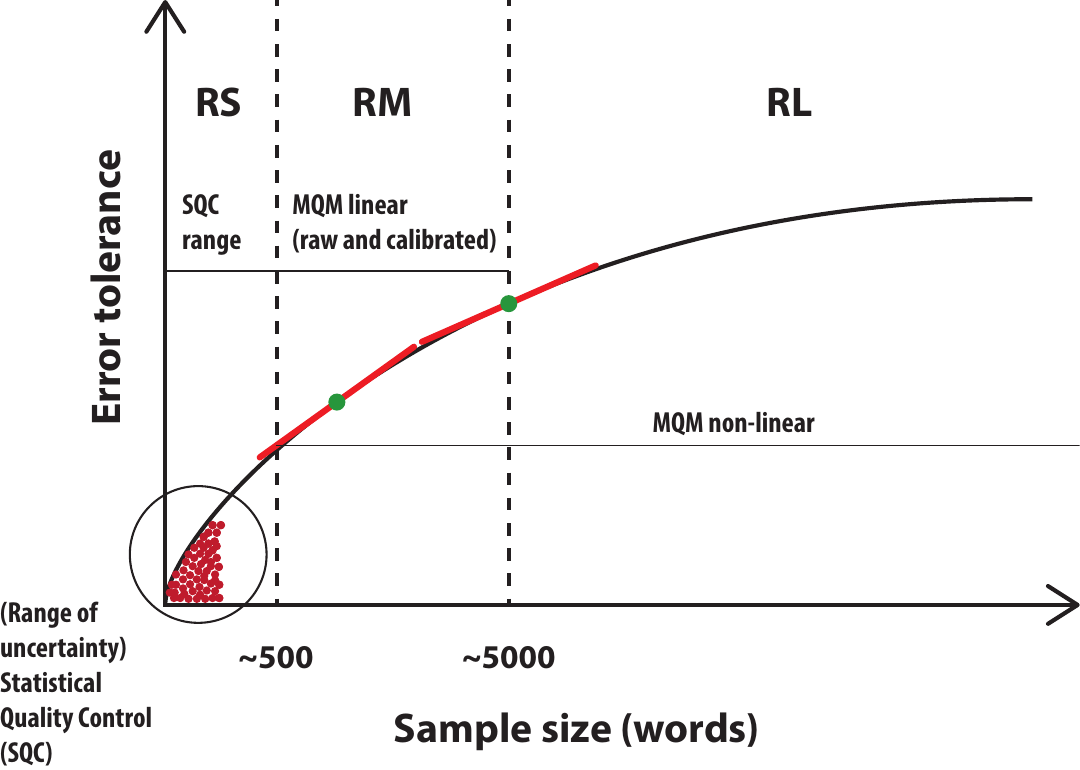} 
    \caption{The chart delineates three distinct sample size ranges—small (RS), medium (RM), and large (RL)—each requiring its own mathematical approach for calculating quality scores. Linear methods can be applied within the small (RS) and medium (RM) ranges but are valid only near the calibration point. In contrast, non-linear methods are applicable across the entire span of the medium (RM) and large (RL) ranges. For the small range (RS), only Statistical Quality Control methods can be used due to the high uncertainty of measurements for very small samples.}
    \label{fig:three-ranges-of-tolerance}
\end{center}
\end{figure*}

As we can see from Figure \ref{fig:linear-vs-nonlinear}, if we calibrate a linear scoring model on 4 pages, it won’t work for 10 pages.
In order for the MQM Metric Scoring Formula to correspond to human judgments and perception for a wider number of sample sizes, we need to use a non-linear scoring model.

The non-linear scorecard is based on standard linear MQM scorecard and uses a logarithmic function to define the score.

\section{Conclusion and Future Work}
\label{sec:discuss}

In this paper we covered a wide scope of sample sizes and different approaches and scoring models. It can be said that this paper represents the Unified Theory of Translation Quality Measurement which explains most use cases of translation quality measurement.
The FULL MQM Error Typology with Calibrated and non-linear scoring is a toolset which allows the reproduction of many different known proprietary metrics.

We have also established the fact that human translation quality evaluation is more than ever THE Golden Standard of measurement and benchmarking for quality measurement, since it is \textbf{the only reliable way to validate any automatic translation quality evaluation}.

    \section*{Acknowledgements}
\label{sec:acknowledgements}
We thank the reviewers for their valuable comments and feedback on our work.
LH and GN are grateful for the grant “Integrating hospital outpatient letters into the healthcare data space” (EP/V047949/1; funder: UKRI/EPSRC).

\begin{small}
\bibliographystyle{apalike}
\bibliography{amta2024}
\end{small}

\appendix



\section{Parameters of the Evaluation Quality Metric}
\label{sec:metric-n-quality-model-app}

\subsection{Calculation Values}

\begin{figure*}[t]
\begin{center}
\includegraphics[width=1\textwidth]{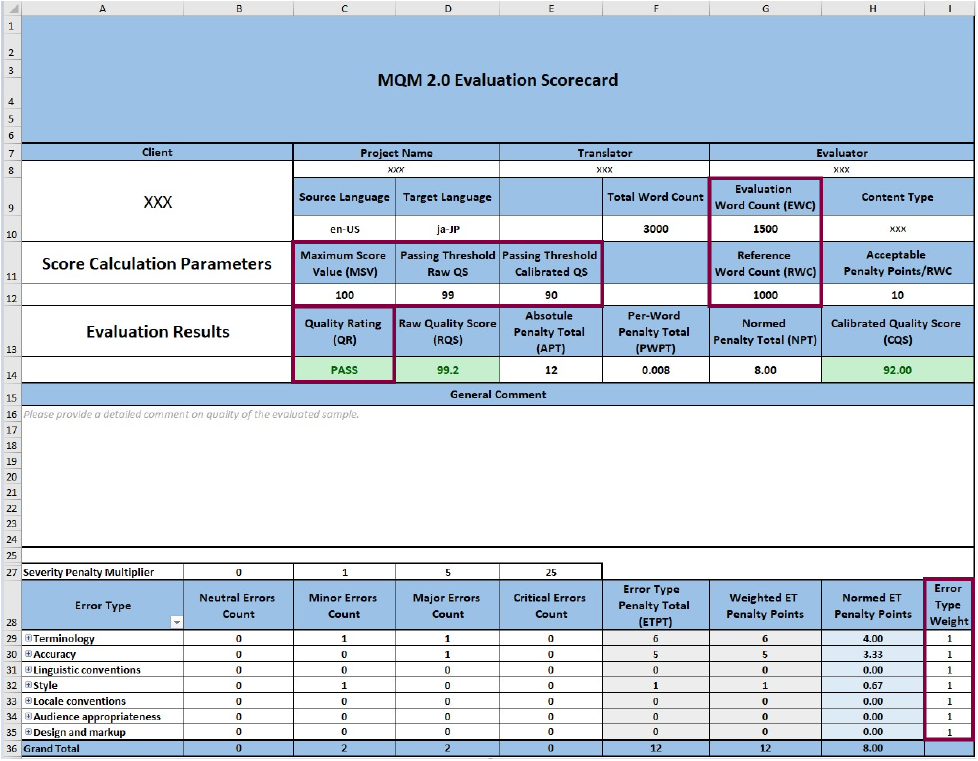} 
    \caption{Sample MQM Scorecard featuring the 7 top-level error dimensions A(29-35) and Severity Penalty Multipliers (Row 27). The most important calculation values are highlighted. }
    \label{fig:mqm2.0-scorecard-important-value}
\end{center}
\end{figure*}

The sample scorecard in Figure \ref{fig:mqm2.0-scorecard-important-value} shows the results of applying one possible metric to a specific evaluation task for a sample of segments (the size of which is stated in the Evaluation Word Count). Rows 29–35 list the selected Error Types -- in this case the seven high-level error dimensions of MQM Core.
The following values highlighted in Figure \ref{fig:mqm2.0-scorecard-important-value} play major roles in designing translation quality evaluation models. The abbreviations listed below are sometimes used when discussing formal equations.

\subsubsection{Evaluation Word Count (EWC)}
The Evaluation Word Count (Figure \ref{fig:mqm2.0-scorecard-important-value}, Cell G10) is the word count of the sample chosen for evaluation. As noted, the EWC can include complete texts, parts thereof, or collections of segments. The EWC is used in the calculation of the Quality Score (QS). The word count according to the draft ASTM standard WK46396 is usually based on the source content.
NOTE: ISO 5060 \footnote{ISO 5060:2024
Translation services Evaluation of translation output
General guidance \url{https://www.iso.org/standard/80701.html}} cites the option to use character counts instead of word counts, or to use line counts that assume uniform characters per line. These approaches accommodate languages that sometimes have dramatically different word counts. ISO 5060 also cites the use of count values for target language content word counts.

\subsubsection{Reference Word Count (RWC)}

The Reference Word Count (Figure \ref{fig:mqm2.0-scorecard-important-value}, Cell G12) is an arbitrary number of words in a hypothetical reference evaluation text. Implementers use this uniform word count to compare results across different projects. The RWC is often set at 1000.

\subsection{Maximum Score Value (MSV)}

The Maximum Score Value of 100 is also an arbitrary value designed to manipulate the Quality Score to shift its value into a range that is easier to understand. It converts the score to a percentage-like value. Cell C12 in Figure \ref{fig:mqm2.0-scorecard-important-value} shows this value for the MSV.

\subsubsection{Passing Threshold (PT)}

The Passing Threshold is the score that defines the Pass/Fail limit. Scoring methods without calibration typically use values such as 0.99 OR 99 – depending on the scale used – as the Passing Threshold (Figure \ref{fig:mqm2.0-scorecard-important-value}, Cell D12). If scoring with calibration is used, the implementer can define any number that is perceived to be visually meaningful, such as 95 or 90 (Figure \ref{fig:mqm2.0-scorecard-important-value}, Cell E12).

\subsubsection{Defined Passing Interval (DPI)}

The Defined Passing Interval is the interval between the Maximum Score Value and the Passing Threshold. In these examples, Raw Scoring Models without calibration use a Defined  Passing Interval of 1 (100-99) or 0.01 (1.00-0.99). When calibrated scores are used, the Defined Passing Interval is magnified to any reasonable range that allows for easy data analysis.

\subsubsection{Final Quality Rating (Pass/Fail)}

The Final Quality Rating (Figure \ref{fig:mqm2.0-scorecard-important-value}, Cell C14) returns a PASS or FAIL quality rating for the evaluated content depending on whether the Quality Score is above or equals the Passing Threshold value (Pass) or is below it (Fail).

\subsubsection{Error Type Weight (ETW)}

Error Type Weights (ETWs) can be used to reflect the importance of Error Types, depending on their importance for a given project, project type or content type. If the ETW is set to 1 for all Error Types (as in the sample scorecard in Figure \ref{fig:mqm2.0-scorecard-important-value}, Cells I (29-35), they are all equally important and result in the same number of penalty points if the Severity Level is the same. If implementers want to distinguish the Error Types by attaching more importance to some Error Types, they can apply different ETWs.

Applying different ETWs can be useful if certain Error Types should be given more prominence than others for a specific type of content. For example, for content with legal implications, implementers may wish to give Accuracy errors higher weight than Style errors. This means that fewer Accuracy errors will be acceptable than Style errors. In other scenarios, a minor Accuracy error may result in fewer penalty points than a minor Style error. For content related to the brands and marketing, implementers can choose to assign higher weights to Style or Audience Appropriateness errors to reflect their importance for this type of content.

\subsection{Error Annotation Values}
\subsubsection{Error Type Number (ET No)}

The sample scorecard shown here reflects Error Type Names assigned to MQM-Core. Optionally, scorecard designers can select other values from MQM-Full or leave out unwanted values. The selected values are listed in the Error Types column (Column B) and associated with Error Type Numbers (ET Nos). Once evaluators have identified a potential translation error, they assign the error instance to one of the Error Types.
\subsubsection{Error Severity Level} 

The Error Severity Level reflects the impact of a particular error on the usability of the text.
Each error instance is annotated according to its Error Severity Level. This sample scorecard features a common set-up with four Severity Levels: Neutral, Minor, Major, and Critical. Three levels, or even two, are also common.
\begin{itemize}
    \item \textbf{Neutral Severity Level}: The neutral severity level is assigned for preferential changes or errors that are not the translator's fault and for which the translator should not be penalized.
    \item \textbf{Minor Severity Level}: Errors that have a limited impact on the usability, understandability or reliability of the content for its intended purpose.
    \item \textbf{Major Severity Level}: The major severity level is assigned to errors that seriously affect the understandability, reliability, or usability of the content for its intended purpose or hinders the proper use of the product or service due to a significant loss or change in meaning or because the error appears in a highly visible or important part of the content.
    \item \textbf{Critical Severity Level}: The critical severity level is assigned to errors that render the entire content unfit for intended purpose or pose a risk of serious physical, financial, or reputational harm. In many quality measurement systems, a single critical error automatically triggers a FAIL rating.
\end{itemize}

\subsubsection{Severity Penalty Multiplier}
\label{subsub:penaltyMultiplier}
The Severity Levels in this sample MQM-based scorecard are represented by Severity Penalty Multipliers. These values can vary depending on implementers’ preferences and needs, but there should be an exponential difference between values for neutral, minor, major, and critical errors. For instance, in this case, the values could be 0, 1, 5, and 25, respectively. This exponential relationship scale reflects the increased risk and impact between the Error Severity Levels. Custom Severity Penalty Multipliers may be required for a variety of reasons: for instance, in case character count per page is used instead of word count.

The Severity Multipliers values times the number of errors at a given Severity Level and the Error Type Weight yields the totals for row values appearing in Figure \ref{fig:mqm2.0-scorecard-important-value}, Cells G (29-35).

\subsection{Scoring Models Parameters}

The scorecard in Figure \ref{fig:mqm2.0-scorecard} comprises the framework for the Raw Scoring Model. The set of framed scoring parameters (defined values and conditions) is used to calculate a Quality Score. This score determines the final Quality Rating (Pass/Fail rating).

\subsubsection{Error Count (EC)}
The Error Count for each error type associated with each Severity Level is multiplied by its respective Severity Multiplier. 
\subsubsection{Error Type Penalty Total (ETPT)}
The Error Type Penalty Total (ETPT) is the sum of penalty points calculated for the individual Error Types annotated in the evaluated text. The error count for a specific Error Type and Severity Level is multiplied by the respective Severity Multiplier and Error Type Weight to obtain the Error Type Penalty Total. 
For example, when using three severity levels, ETPT is defined as ((Minor Error Type count × Minor Severity Multiplier) + (Major Error Type count × Major Severity Multiplier) + (Critical Error Type Count × Critical Severity Multiplier)) × Error Type Weight. 
\subsubsection{Absolute Penalty Total (APT)}
The Absolute Penalty Total is the sum of all Error Type Penalty Totals (Figure \ref{fig:mqm2.0-scorecard}, Cell E12). APT is the most important value used for Quality Score calculation.
\subsubsection{Per-Word Penalty Total (PWPT)}
The Per-Word Penalty Total (Figure \ref{fig:mqm2.0-scorecard}, Cell F12) is determined by dividing the Absolute Penalty Total by the Evaluation Word Count. The Per-Word Penalty Total is also one of the key values that contributes to the Raw Quality Score calculation.
\subsubsection{Normed Penalty Total (NPT)}
The Normed Penalty Total (Figure \ref{fig:mqm2.0-scorecard}, Cell G12) represents the Per-Word Error Penalty total relative to the Reference Word Count.
Typically, 1000 is used as the arbitrary number to represent the Reference Word Count; therefore NPT is sometimes referred to as the Error Penalty Total per Thousand Words.
The Normed Penalty Total is obtained by multiplying the PWPT by RWC (NPT = PWPT × 1000 in our example). This is mathematically equivalent to (APT × RWC)/EWC.

\subsubsection{Quality Score (QS)}
The Quality Score is the primary quality measure of a translation product.

\subsection{Calculating the Linear Quality Score}
There are two ways to calculate the Linear Quality Score: with and without calibration.

\subsubsection{Quality Score without Calibration (Raw Score)}
The Raw Linear score determines the portion of the text containing errors, subtracts this number from 100, and thus provides a value representing the error-free section of the evaluated sample.

Logically then, the Quality Score expresses the portion of the evaluated target content that is correct. In this example, the acceptable interval set as allowed for the “portion with errors” is 1. Hence, any quality score between 100–99 (1–0.99 respectively) produces the Pass rating.

The acceptable interval is delimited by the Acceptable Penalty Points (APP) value for the Reference Word Count, which corresponds to the Passing Threshold. For example, a requester of legal translation might find that their Passing Threshold would be a Raw Score of 99.5 (e.g., five penalty points for a thousand-word Reference Word Count), while a requester for user-to-user technical help might accept a raw score of 97.2 (e.g., 28 penalty points for the same Reference Word Count).

However, relying on Raw Score calculations alone has drawbacks. For the legal example, the score hovers too close to 100, making it difficult to use the Raw Scores. In addition, if an organization has multiple content types, each with their own Passing Threshold, it can be difficult to track and apply the proper threshold to each one. Setting an acceptance threshold using Raw Scores is challenging when varying scores end up looking very close to each other, as such acceptance thresholds are not necessarily intuitive. The threshold may even turn out to be a complex fractional value, which means that simply scaling the Raw Score does not solve this problem.

\subsubsection{Quality Score with Calibration}
The second option is to calibrate the penalty points calculated for the evaluated sample against a preselected Passing Threshold or tolerance limit on a special calibrated quality scale. Calibration expresses the scoring values in a way that stakeholders can interpret easily in line with their expectations and specifications. 

To do so, implementers specify an Acceptable Penalty Points (APP) value during the project specification stage, representing how many penalty points they would deem acceptable for the Reference Word Count on a calibrated quality score scale. They then associate this tolerance limit with the Passing Threshold.

In its raw form, a score is initially calculated as described in the previous section. It is then converted to a Calibrated Score scale by scaling the raw passing interval to Calibrated Passing Interval and mapping the raw score to Calibrated score on Calibrated Score scale, as shown below.

Calibration applies the aforementioned ergonomic Passing Threshold. This Passing Threshold differs from the Raw Quality Score. For example, the Defined Passing Interval in Figure 5 is 100-85, where 85 will be a PASS and anything less will be a FAIL. In this case the 85 Passing Threshold corresponds to the maximum acceptable number of errors on a Raw Score scale (for example, the five penalty points for the legal translation or 28 for the technical help example).

The calibration process acts like a magnifying glass for viewing the otherwise very small or inconsistent acceptance ranges close to 100. This approach makes the quality rating easier to use and understand, highlighting differences in translation quality for evaluated texts more clearly.

\subsection{Score Calculations}
\label{subsec:calibration}
\subsubsection{Calculating the raw quality scores}
Calculating scores without calibration uses the steps shown in Figure \ref{fig:mqm2.0-scorecard-formulas}. See the Appendix: Scoring Model Parameters for a list of all parameters and their abbreviations.

\subsubsection{Calculating the quality scores with calibration}
The Scoring Method with Calibration enables implementers to account for the error tolerance for a specific word count (Reference Word Count) and to link it to the pre-defined Passing Threshold (PT), against which the Pass rating is determined. For a list of all parameters and their abbreviations, see the Appendix: Scoring Model Parameters.

The scoring formula for calculating the quality score with calibration works with the standard calculation values, such as Evaluation Word Count, Absolute Penalty Total and Normed Penalty Total. However, a few additional values and parameters have to be defined. These are used to pre-define the specified acceptance criteria (the error tolerance) and to link these criteria to a scale that should be understandable or appropriate for all stakeholders.
The following values used in the score calculation above are pivotal for a score calibrated with respect to a predefined Passing Threshold.

\textbf{Acceptable Penalty Points (APP) for the Reference Word Count}
Penalty points are deemed as still acceptable for a certain volume of text, typically for the Reference Word Count of 1000 words.
Typical questions to ask when defining the Acceptable Penalty Points are:
\begin{itemize}
    \item What is the number of Minor errors that would still be a Pass for a sample of 1000 words?
    \item What is the number of Major errors that would still be a Pass for a sample of 1000 words?
\end{itemize}
In simple terms, the APP reflects the error count that stakeholders would still consider to be acceptable for a given word count (typically 1000 words) provided that the Minor Error Weight is 1.

In the current example, the acceptable error tolerance is defined as 10 minor errors OR 2 major errors per 1000 words, which yields a Raw Quality Score of 99. If the Normed Penalty Total calculated for the evaluation sample is greater than 10 penalty points, the defined Passing Threshold has been exceeded and the evaluation result is FAIL.

\textbf{Passing Threshold (PT)}
A number perceived as an intuitively reasonable Passing Threshold.
Calibration enables the determination of a Passing Threshold that is psychologically meaningful to stakeholders. This number typically is any reasonable number in the range of 0-100. It represents the Passing Threshold score that is linked to the pre-set count of penalty points for the reference word count, i.e. the initially defined error tolerance for a certain unit of text.
Calibration transforms the narrow passing interval obtained using the raw, uncalibrated score to a wider and more interpretable interval, which acts analogous to a magnifying glass.

\textbf{Scaling Factor (SF)}
Parameter to scale the Acceptable Penalty Points (APP) for the reference word count across the Defined Passing Interval (DPI).
Let's consider Figure \ref{mqm2.0-scorecard-scaling-factor}. On the top Raw Scale the raw passing threshold is 98, which means that a maximum of 20 raw penalty points are allowed on a sample of 1000 words. On a Calibrated Score scale (bottom) the Defined Passing Interval (DPI) is 15 (upward from 85 to 100). Therefore, the raw Passing Interval scales \textbf{\textit{down}} from 20 on the raw scale to 15 on the Calibrated Scale. The Scaling Parameter for this new value will be 15/20=0.75.
On the Raw Scale, the raw score is 98.44, which means that NPT=15.6 (the error density of a sample is equivalent to 15.6 errors on 1000 words).
The trick to the Calibration is that what is scaled is not a passing interval, but rather, NPT. We multiply raw NPT of 15.6 by a scaling factor 15/20=0.75, which resolves  in NPT = 11.7 on the calibrated scale. Therefore, the calibrated score will be 100-11.7=88.3 (bottom scale).

\section{Further Discussion}

As shown herein, the Translation Quality Score calculation depends and is a function of many conditions and parameters:

\begin{itemize}
    \item Client specifications, defining tolerances for various content types, purposes, and audiences.
    \item Language pair, culture.
    \item Purpose of evaluation.
    \item Measurement conditions and requirements.
    \item Sample sizes.
    \item Technology platforms used (MT, AI, TMS, etc.).
\end{itemize}

Our further research directions include:

\begin{itemize}
    \item Developing practical methods of reliable and simple translation quality score calculations for smaller samples using Statistical Quality Control methods.
    \item Developing standardized score cards for various use cases, with examples.
    \item Improving reliability of automatic GenAI-enabled quality measurement methods.
    \item Benchmarking more annotation and quality evaluation data to develop and provide ways to validate automated quality evaluation metrics.
\end{itemize}

\section{Highlighted Scorecards and Model Parameters}
\label{sec:appendix}

We list the sample MQM scorecard figures here with different highlights mentioned in the paper.

\begin{figure*}[t]
\begin{center}
\includegraphics[width=1\textwidth]{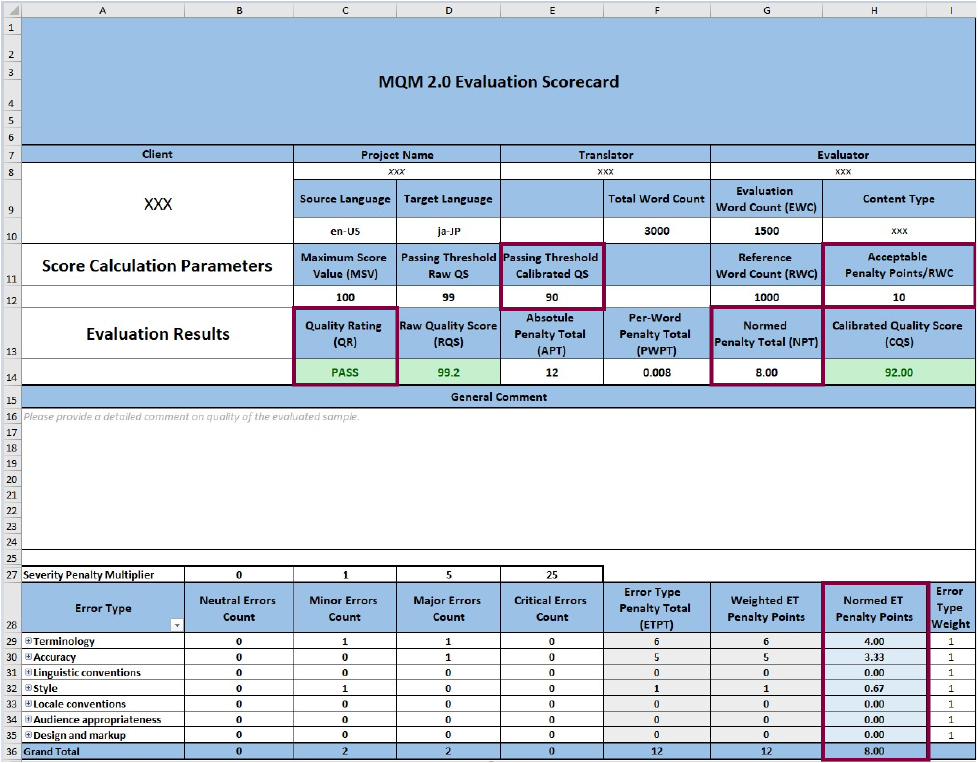} 
    \caption{In this example, 90 is defined as the Passing Threshold. This score is achieved if the Normed Penalty Total for the Evaluation Word Count is 10. If it is greater than this value, the Quality Score is a number below the Passing Threshold of 90, and the evaluation result is FAIL. }
    \label{fig:mqm2.0-scorecard-passing-threshold}
\end{center}
\end{figure*}

\begin{figure*}[t]
\begin{center}
\includegraphics[width=1\textwidth]{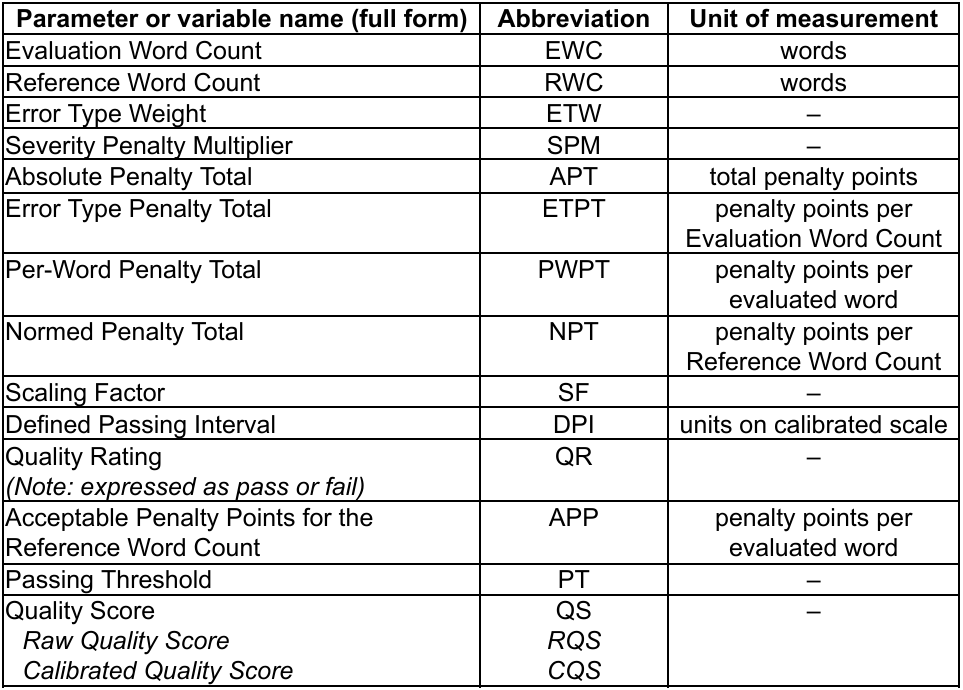} 
    \caption{Scoring Model Parameters and Variables}
    \label{fig:mqm2.0-scorecard-parameters}
\end{center}
\end{figure*}

Scoring Model Parameters and Variables are explained in Figure \ref{fig:mqm2.0-scorecard-parameters}.

\section{The MQM Metric Deployment Process (use case)}

A typical use case of MQM deployment often includes analysis of previously collected evaluation results, which are used to validate new would-be deployed MQM Scoring Model against established practices of quality tolerance thresholds and specifications.

Here’s the typical use case:
\begin{itemize}
    \item Before we introduced our new MQM-based error typology with the weights and multipliers and thresholds, we analyzed previously conducted evaluations. We had already been marking errors, but the decision on whether the translation was acceptable or unacceptable was left to the evaluator, irrespective of errors. So, we could have an evaluation with a two or three errors that was rated unacceptable, and evaluations having twenty errors that turned out to be acceptable. In principle, marking errors was just for educational purposes, and the decision on acceptability was a holistic one. So we took all this data and calculated on average how many errors marked the threshold for unacceptable translations. We then prepared several options for weights and multipliers and played through with them to see what comes the closest to the identified threshold. Then we took the chosen weights and multipliers and tested them on actual live translations. We always ask the evaluators to mark if the score and acceptability corresponded to their actual feelings about the translation. This is how we established our current MQM-based new methodology.
\end{itemize}

\end{multicols}
\end{document}